Choose Your Own Adventure:

Interactive E-Books to Improve Word Knowledge and Comprehension Skills

Stephanie L. Day[1]

Jin K. Hwang[2]

Tracy Arner[3]

Danielle McNamara[3]

Carol M. Connor[2]

[1]University of Central Florida

[2]University of California, Irvine

[3]Arizona State University

Corresponding Author: Stephanie Day; stephanie.day@ucf.edu, 4000 Central Florida Blvd, P.O. BOX 162993, Orlando, FL 32816-2993

This work was supported by grants from the U.S. Department of Education, Institute of Education Sciences, Developing Electronic-Books to Build Elementary Students' Word Knowledge, Comprehension Monitoring, and Reading Comprehension (Grant number R305A170163), and Reading for Understanding Network (Grant number R305N160050).

Declarations of interest: none

**Abstract**

The purpose of this feasibility study was to examine the potential impact of reading digital interactive e-books, Word Knowledge e-books (WKe-Books), on essential skills that support reading comprehension with third-fifth grade students. Students ($N = 425$) read two WKe-Books, that taught word learning and comprehension monitoring strategies in the service of learning difficult vocabulary and targeted science concepts about hurricanes. We investigated whether specific comprehension strategies –1) word learning and strategies that supported general reading comprehension , 2) summarization, and 3) question generation, show promise of effectiveness in building vocabulary knowledge and comprehension skills in the WKe-Books. Students were assigned to read one of three versions of each of the two WKe-Books, each version implemented one strategy. The books employed a choose-your-adventure format with embedded comprehension questions that provided students with immediate feedback on their responses. Paired samples t-tests were run to examine pre-to-post differences in learning the targeted vocabulary and science concepts taught in both WKe-Books. For both WKe-Books, students demonstrated significant gains in word learning and on the targeted hurricane concepts. Additionally, Hierarchical Linear Modeling (HLM) revealed that no one strategy was more associated with larger gains than the other. Performance on the embedded questions in the books was also associated with greater posttest outcomes for both WKe-Books. These findings suggest that the affordances offered by technology, which are unavailable in paper-based books, can effectively support students' development of reading-related skills, including strategy use. Further, this work discusses important considerations for implementation and future development of e-books that can enhance student engagement and improve reading comprehension.

1. Introduction

Rapid advances in the development of mobile technologies have increased the use of various portable devices in schools intended to engage students and improve learning outcomes (Bedesem & Arner, 2019; López-Escribano et al., 2021). Among them, electronic-books (e-books) are read frequently by school-aged students on Kindles, tablets, and/or laptops and have the potential to increase students' engagement while reading a variety of texts (Swanson et al., 2020; Xu et al., 2020). The number of e-books used in schools is expected to continue to increase, and thus, the appropriate and relevant use of electronic resources have become significant issues regarding promoting and maintaining students' learning outcomes. While some studies have reported a "screen inferiority" effect of reading on digital devices (Clinton, 2019; Delgado et al., 2018), an increasing number of studies also report that some features in digital books effectively promote vocabulary learning and comprehension skills (Korat et al., 2017). A recent meta-analysis examined the impact of print books versus digital books on reading outcomes across 39 studies. Results indicated that enhancements offered by digital books, such as features that focused on reviewing story content and encouraging comprehension strategy use (e.g., retrieving background knowledge), led to improved learning compared to print books (Furenes et al., 2021).

The purpose of the present feasibility study was to expand upon a previous randomized control trial (RCT) in which the use of an interactive Word Knowledge E-Book (WKe-Book) significantly boosted students' word learning and comprehension. In this study, we aimed to assess the feasibility of moving beyond a focus of just teaching words to teaching concepts through deeper text-meaning comprehension strategies to support students' acquisition of knowledge and practice of specific reading skills by enhancing the original WKe-Book and developing an additional WKe-Book.

**E-Books and Instruction**

The integration of e-books as instructional tools can support student learning both in and out of the classroom (Korat et al., 2017). Regardless of the device used to access e-books, there are several features that can enhance or support students' reading experience in ways that a static, paper book cannot. The affordances of the dynamic technology can support students' acquisition of literacy skills through improved accessibility, providing a more engaging and personalized learning experience, and targeted literacy support (Connor et al., 2019; Kaynar et al., 2020). In addition to improving reading experiences, e-books can also provide valuable insights into students' reading skill development through user-log and teacher report features to track student progress (Huang & Liang, 2015).

Decades of research indicate that students' literacy skills benefit from reading more. The more students read, the better they get at reading and students who are better readers tend to enjoy reading more (Willingham, 2017). Thus, students who struggle, tend to experience lower engagement with reading materials of any kind. A primary benefit of using e-books as an instructional tool is the increased engagement stemming from the dynamic technology. For example, including multimedia effects such as video, images, and sound effects can increase students' interest and engagement with the book (Reich et al., 2016).

E-books can also support comprehension beyond increasing engagement by including features that help bridge gaps in students' knowledge or reading skill (Korat et al., 2021). For example, some e-books include text-to-speech features that allow students to have the text 'read aloud' when they encounter unfamiliar words. Similarly, many e-books include features that support vocabulary acquisition, such as, highlighting important words or 'built-in' definitions accessible by tapping the word (Korat et al., 2022; Swanson et al., 2020).

The two e-books tested in this feasibility study, one narrative and one hybrid (i.e., narrative and informational) text, were designed with affordances to support students' acquisition of word knowledge and comprehension strategy use to improve their reading comprehension.

**Development of WKe-Books**

The WKe-Books were developed iteratively by embedding reading comprehension strategies to support third-fifth grade students to read for understanding (Connor et al., 2019; Umarji et al., 2021; Yang et al., 2021). The WKe-Books were designed to engage young readers through an interactive design using a choose-your-own-adventure and name-your-character format, while also supporting comprehension skills. The development focus in the initial WKe-Book was on general comprehension strategies with an emphasis on word-meaning learning strategies such as contextual analysis (e.g., teaching students how to use context clues to infer the meaning of unfamiliar words), morphemic analysis of words (e.g., teaching students how to divide words into meaningful parts to infer meaning), and dictionary use (e.g., encouraging students to look up unfamiliar words). These strategies were chosen based on previous research on reading for understanding (Korat et al., 2013). They were embedded in the initial WKe-Book, *Dragon's Lair*, and were explicitly taught and used during small groups (Connor et al., 2019).

Previous studies using the *Dragon's Lair* WKe-Book have shown positive results in improving students' reading and metacognitive outcomes (Connor et al., 2019; Umarji et al., 2021; Yang et al., 2021). Specifically, Connor and colleagues (2019) found significant treatment effects of the WKe-Book on third through fifth-grade students' word-related outcomes in their RCT. Students in that RCT were first randomized to be in the WKe-Book or no WKe-Book condition. The students in the WKe-Book condition were randomized again to be assigned in two different treatment conditions: WKe-Book with book club or without book club. In both treatment conditions, all students read the WKe-Book three times a week for three weeks. In the no book club condition, students read the WKe-Book independently and did not take part in book club review sessions. Students in the book club condition met with the researchers once a week for 15 minutes, during which time word learning strategies (i.e., morphemic analysis, contextual analysis, and dictionary use) were explicitly taught in groups of 5-6 students. Connor and colleagues (2019) found that reading the WKe-Book helped improve students' word knowledge, word knowledge calibration, and word-learning strategy use. This, in turn, also predicted students' scores on reading comprehension and vocabulary measures. Students who also participated in weekly book clubs had even greater increases than those who only read the WKe-Books independently. Additionally, when tested after a six-week delay, students retained their knowledge of the complex vocabulary taught in the story. Results from this study showed that technological affordances of interactive e-books can help promote students' reading-related outcomes and this effect can be moderated by small group instruction.

      The goal of this feasibility study was to enhance the previous version of the WKe-Book and to develop a new WKe-Book by adding new deeper, text-meaning strategies and examining whether they show promise in promoting third-fifth grade students' reading-related skills. To that end, we added text-meaning strategies including summarization and deep-level question generation to the word learning strategies that supported general comprehension taught in the original WKe-Book to assess potential differences in learning outcomes. Summarization and deep-level question generation were added because they are malleable comprehension repair strategies commonly taught to readers in elementary and middle school to develop higher-level

inferencing from different types of text (McNamara, 2007). Although the aim of *Dragon's Lair* was on word-meaning comprehension, as text is composed of words, text-meaning strategies like summarization and question generation could also enhance word-meaning comprehension. As we aimed to assess the feasibility of teaching text-meaning strategies, the second WKe-Book, *Hurricane!* was designed to learn scientific concepts about hurricanes, thus moved beyond just word-meaning comprehension.

### Summarization

Summarization requires students to identify the main ideas and purpose within a text, while dismissing information that is irrelevant or redundant (Brown et al., 1981). This strategy is important for learning because it encourages students to attend to higher-level meanings of texts, consequently increasing comprehension (Bransford et al., 2000). When students are planning to summarize information in texts, they are more likely to read intentionally for understanding. Accordingly, students who summarize are better able to monitor and repair their own comprehension of information in a text (McNamara et al., 2011). According to Wade-Stein and Kintsch (2004), summarizing text promotes students' construction of factual knowledge and conceptual knowledge because the process of summarizing reinforces students' memory representations of its content. Taken together, prior research suggests that the summarization strategy increases students' retention of text material, particularly for lower-achieving students (Gil et al., 2010).

### Deep-Level Question Generation

Generating deep-level questions encourages students to question the meaning of text and their understanding of text while they read. Evidence indicates this technique is effective because it promotes active engagement with text and encourages students to identify gaps in their own understanding, and subsequently repair those gaps (Rosenshine et al., 1996). Additionally, students are more likely to demonstrate long-lasting understanding of a text when they link information from the text to their own prior knowledge while generating questions (Caillies et al., 2002).

In the current study, classrooms were randomly assigned (counterbalanced within each district and grade) to read one of three versions of each WKe-Book. For *Dragon's Lair*, the strategy versions included: word learning (e.g., contextual and morphemic analysis), summarization, and deep-level question generation. For *Hurricane*, the strategy versions included: general comprehension, summarization, and deep-level question generation. The general comprehension strategy in *Hurricane!* was designed to parallel the word learning strategy version of *Dragon's Lair*, including word learning strategies for unfamiliar vocabulary and strategies to support recall such as rereading. Throughout the WKe-Books, multiple-choice questions for each targeted strategy were embedded in the text and students were provided with explicit feedback on how to use the targeted strategy to help them answer each of them. Daily review sessions led by an interventionist included explicit instruction on how to employ the targeted strategies and a story discussion.

**Study Aims**

In this feasibility study, we build upon the previous study by 1) embedding text-meaning strategies including question generation and summarization strategies in addition to word learning strategies in the WKe-Books to examine whether these strategies were associated with third-fifth grade students' learning outcomes and 2) developing an additional hybrid WKe-Book, *Hurricane!*, in which students were taught how to employ one of three targeted comprehension strategies in the service of learning science concepts about hurricanes. To this end, students read one or two WKe-Books (i.e., *Dragon's Lair* and *Hurricane!*), which were presented in one of three versions that differed in the focal strategy: word learning strategies in *Dragon's Lair* (i.e., morphemic and contextual analysis and dictionary use), general comprehension strategies in *Hurricane!,* and summarization and deep-level question generation in both books**.**

1. For each of the WKe-Books, we posed the following questions to assess promise of effectiveness:
    a. Did students make significant posttest gains in word knowledge for *Dragon's Lair* and hurricane concepts knowledge in *Hurricane!*? Further, were there differences in posttest gains as a function of strategy version (i.e., word learning, general comprehension, summarization, and question generation) that students read for each WKe-Book?
    b. Were students' WKe-Book version specific knowledge gains dependent upon their baseline vocabulary, decoding, and comprehension skills?
    c. Was higher performance on the embedded comprehension questions associated with greater posttest outcomes in word knowledge or hurricane concepts? Did students who read the books more than one time exhibit greater posttest gains?
2. What were teachers' perceptions of the feasibility of using the WKe-Books as an instructional tool for improving comprehension skills?

## 2. Methods

*Participants*

Four hundred and twenty-one students in third through fifth grade and 16 teachers, in two public school districts in Southern California and Central Arizona participated in the study. One school in Arizona (AZ) participated in this study; 68% of the students at the school qualified for free or reduced-price lunch. Seventy-one percent of the students at the school were of Hispanic descent, 13% White, 11% Black, and 5% other ethnicities. Three schools in California (CA) participated from a district where, 37% of the students were White, 36% Asian, 20% Hispanic, and 7% other or mixed ethnicities. Approximately 32% of these students qualified for free or reduced-price lunch. All 421 participants read *Dragon's Lair*, while only the CA sample ($n = 271$) also read *Hurricane!*.

*The WKe-Books*

Two WKe-Books were developed and used in this study. *Dragon's Lair* (McDonald, 2012) was expanded from a story written by a professional children's author and tells the story of a young boy (named by the reader) and his magic unicorn, Finn, who go on adventures together. In *Dragon's Lair*, Finn and the boy travel to a village where evil green dragons are kidnapping children and taking them into the caves to work for them. The young boy meets a girl (named by the reader) in a shelter during a green dragon raid. The children and Finn have a series of adventures where they find the caves, save kidnapped twin brothers, meet the good red dragons, work with them to defeat the green dragons, and secure a magical emerald. The six-chapter story includes multiple decision points and one to two comprehension questions for each chapter with a total of 5 to 9 questions, depending on the chosen story stream. Due to the total number of questions being unbalanced, we left scores on the embedded questions out of the current analysis. Target vocabulary words were selected using the Academic Word List (Coxhead, 2000), SAT/ACT Word List, and Spanish cognates. The average number of words was 994 words per chapter and the average Flesch-Kincaid Grade Level was 3.8. Readability metrics and screenshots from *Dragon's Lair* can be found in Umarji et al. (2021).

The newly designed WKe-Book was a hybrid science book titled *Hurricane!* (Connor et al., 2020). In *Hurricane!,* Finn and the young boy (named by the reader) are whisked away to an island threatened by an approaching hurricane. They meet two sisters (named by the reader) who share that their dog is lost. Together, the group must search for the dog while making careful decisions to stay safe from the hurricane. *Hurricane!* also included comprehension strategies, but differed from *Dragon's Lair* by adding five target concepts about hurricanes: (1) What is a hurricane and why is it dangerous? (2) How does a hurricane form and what are the parts of a hurricane? (3) What is a storm surge and why is it dangerous? (4) How do meteorologists study and track hurricanes? and (5) How do we stay safe in hurricanes? The structure of *Hurricane!* was similar to *Dragon's Lair* in that there were multiple story streams and decision points, but it had only four chapters (Appendix). The average word count in each chapter was 2,134 words and the average Flesch-Kincaid Grade Level was 4.16. Regardless of the story stream students chose, all five target concepts were presented with corresponding comprehension questions in each chapter for a total of nine questions in the text. At the beginning of the story, all students were presented with an embedded short video that demonstrated how hurricanes form in the ocean. One or two illustrations were included in each chapter. While some were representative of the fictional part of the story, others were used to illustrate the hurricane concepts, such as how hurricanes form (see Appendix).

Sample questions from each strategy version for both books are provided in the Appendix. Three versions were created for both *Dragon's Lair* and *Hurricane!;* each focused on one of the comprehension strategies. For *Dragon's Lair,* the versions included: word learning, summarization, and question generation. While we aimed to create a comparable version of the lower-level word-meaning strategies in *Dragon's Lair* to compare with *Hurricane!*, since this book was focused on concepts, we also taught students lower-level text-meaning strategies such as rereading to support story recall. We refer to this strategy as a general comprehension strategy. The other two versions of Hurricane also included summarization and question generation as used in *Dragon's Lair*.

For the word learning strategies taught in *Dragon's Lair* and the general comprehension of the concepts taught in *Hurricane!* versions, we did not include open-ended questions. In the question generation and summarization versions of both WKe-Books, students were presented with both multiple-choice questions and open-ended questions, however analysis of the open-ended responses was beyond the scope of this initial feasibility study. Children received feedback and were asked to try again when choosing incorrect responses for multiple-choice questions in all versions. The embedded questions in each book focused on asking questions about the vocabulary in *Dragon's Lair* and the targeted concepts in *Hurricane!*. Although *Dragon's Lair* was more focused on learning vocabulary, even the summary and question generation versions focused on encouraging students to include the targeted words. For example,"What would be the best question to ask yourself right now?", answer choices would include the vocabulary (i.e., "Why did Charlie act so impulsively?") and feedback would encourage students to think about whether they understood the meaning of the word (see Appendix).

*Measures*

**Word Match Game** (WMG; Connor et al., 2022). The WMG is an online, adaptive assessment created to measure word knowledge and document change over time for kindergarten–fifth grade students. Students receive auditory instructions, prompting them to select two paired words from a set of three words (e.g., cat, tree, and kitten). Item response theory (IRT) analysis on the item pool ($n > 200$) indicates that item difficulty ranges from -3.3 to 3.5, the overall test information is excellent with a bell-shaped information function, and total test information greater than 2.0. Growth scaled scores were used in the analyses. We were only able to administer the WMG at pretest due to the COVID-19 pandemic.

**Letters2Meaning** (L2M; Connor et al., 2022). L2M is an online, adaptive assessment that measures decoding, letter-word reading, spelling, and comprehension for kindergarten–fifth grade students across six different subtests. Letter, Sound, and Word Identification presents words aurally and the student selects the letter or word from among four to six choices. The Letters2Words subtest measures students' ability to choose letters from a bank and sequence them to create a given word by manipulating them on the screen (e.g., rearrange the letters to spell the word cat). The Words2Sentences subtest asks students to sequence/rearrange words from a bank to build a sentence. The IRT analysis on this assessment indicates an item difficulty range of -6.5 to +9.3, overall test information is excellent with a bell-shaped information function, and the total test information is greater than 2.0 throughout the range of -5.0 to +5.0. Growth scaled scores were used in the analyses. Due to school closures, we were only able to administer L2M at pretest.

**Dragon's Lair Word Knowledge Task**. The Word Knowledge Task was a paper-based, whole-class assessment based on the target vocabulary (25 words total) introduced in *Dragon's Lair*. Thus, it was intended to measure the type and number of words learned after reading the WKe-Book. The three-part assessment included Matching Vocabulary, What's the Meaning of This?, and Let's Figure it Out. Matching Vocabulary required students to match the word with the definition. What's the Meaning of This? asked students to read a sentence and choose a synonym for a target word from a bank of three words. Let's Figure it Out required students to read a sentence and then write the definition of an underlined target word (worth up to 3 points each). There were a total of 25 questions on the entire assessment with 10 questions each for the first two sections and 5 questions for Let's Figure it Out, with 35 total points possible. Previous studies using this measure suggest adequate reliability ($\alpha = .70$) and sensitivity to word knowledge gains. Students completed the Word Knowledge task just before and right after reading *Dragon's Lair*.

*Hurricane!* **Concept Knowledge Task.** The *Hurricane!* Concept Knowledge Task was developed specifically for the *Hurricane!* WKe-Book to measure whether students learned the five target concepts of the books. The test consists of 10 items, totaling up to 19 points. The first six items are multiple-choice items worth 1 point for a correct answer and 0 for incorrect answers. Items 7-10 are open-ended questions. Items 7 and 8 are open-ended questions asking students to write how to stay safe in a hurricane and what is storm surge. Item 7 is worth up to 4 points and Item 8 is worth up to 3 points. Item 9 asked students to generate a question about how hurricanes form in the ocean and is worth up to 2 points. Item 10 asks students to write a summary of how hurricanes form and is worth up to 4 points. Research assistants were trained to score student responses. Inter-rater reliability ($\kappa = .87$) was acceptable. Reliability ($\alpha = .66$) for the posttest was adequate. The test is paper-based and was completed independently and in large groups.

**Student user log data.** *Dragon's Lair* and *Hurricane!* student user-log data recorded by the WKe-Book software was used to assess responses to the embedded comprehension questions and to interrogate the number of times students read the WKe-Books. Total scores for the embedded multiple-choice questions were calculated for each student based on the number of attempts required to achieve a correct response for each question. Scores for *Hurricane!* were calculated as follows: 4 points for answering the question correctly the first time, 3 points on the second response, 2 points on the third response, 1 point for the fourth response, and 0 points for taking more than 4 tries to answer correctly. If students completed the first reading of each book early, they were encouraged to read the story again, choosing a different story path. Thus, we also calculated a total read count to represent how many times students read each of the books.

**Procedure**

Students were first administered the online assessments (WMG and L2M) and the Word Knowledge Task (pretest) for *Dragon's Lair*. All students then read *Dragon's Lair* for 20 minutes per reading session, three days a week during the fall semester of the school year. As the current study was intended to be a design and feasibility study (and for ease of implementation), we opted to deliver whole group book club review sessions. For the first 10 minutes of each session before students read, trained interventionists reviewed the target strategy assigned to each class. Classroom teachers helped ensure that their students remained on task during each session. Sample lesson plans are available upon request from the authors. *Dragon's Lair* took students approximately nine reading sessions to complete. Once each class had finished reading *Dragon's Lair*, students were administered the Word Knowledge Task. The CA students then moved on to read *Hurricane!* in the late winter. The procedure for reading *Hurricane!* was the same as *Dragon's Lair.* Students read the second WKe-Book in six sessions, after which they completed the *Hurricane!* Knowledge Task.

Classrooms were randomly assigned to read either the word learning strategy of *Dragon's Lair*, summarization, or deep-level question generation version of each book. All students (teachers/classrooms = 16) read *Dragon's Lair* first. Six classrooms read the word learning version and the summarization version, and seven read the question generation version.

Versions were counterbalanced by grade and district. For example, if there were three third grade teachers in one district, they were each randomly assigned one of the three versions. Nine classrooms in the CA district read *Hurricane!*. Versions were again counterbalanced within grade level. There were three classrooms at each grade level so one class at each grade level was semi-randomly assigned to one of the three versions (general comprehension, summarization, or question generation) such that no students were in the same condition as when they read *Dragon's Lair*. For instance, if students read the summarization version of *Dragon's Lair*, they were assigned to read either the general comprehension or deep-level question generation version for *Hurricane!*.

The study plan in the spring of 2020 was impacted by school closures due to the COVID-19 pandemic. Therefore, AZ students were not able to read *Hurricane!*. and the CA students were unable to complete the WMG and L2M assessments at posttest. Finally, we were unable to conduct exit interviews with students as schools closed abruptly.

### 3. Results

Descriptive statistics for all measures are provided in Table 1. For each of the WKe-Books (i.e., *Dragon's Lair*, *Hurricane!*), we assessed differences in posttest gains as a function of strategy version, potential differential impacts as a function of baseline vocabulary and decoding and comprehension skills (i.e., WMG and L2M pretest scores), and the association between comprehension question performance and posttest outcomes. A paired samples *t*-test was run to assess differences in pre to posttest performance. Hierarchical linear modeling (HLM, Raudenbush & Bryk, 2002) was used for the remaining analyses to account for the nested structure of the data – students nested within teachers.

***Dragon's Lair.*** To determine whether students had significant gains between the *Dragon's Lair* Word Knowledge Task pretest and posttest, we first ran a paired samples *t*-test which was significant (*t*(420) = 29.702, *p* < .001). Students' posttest scores on the *Dragon's Lair* Word Knowledge Task were then entered as the dependent variable in a hierarchical two-level model in HLM (Table 2). Pretest scores were added to the model. Students' performance on both WMG and L2M were included as covariates to control for their baseline language and literacy skills. Grade level was entered as a teacher/classroom-level covariate in the model and was centered at fourth grade.

Overall, higher pretest and L2M (decoding and comprehension) scores were associated with higher scores on the Word Knowledge posttest. We also found evidence of a grade effect such that on average, students in fourth and fifth grade tended to perform better than the third-grade students on the posttest (Table 2).

To determine whether there were differences by strategy version students read, we created three dummy variables for each version of the book, coded as 1 or 0. For example, if one class read the word learning version of the book, they received a 1 and the other two groups were coded as a 0. In each of the three strategy models, posttest scores on the Word Knowledge Task were entered as the dependent variable. The model included the summarization and question generation dummy variables (with word learning serving as the reference group). The summary and question generation versions were not determined to have greater post-test outcomes compared to the word learning version of the book as was tested in the WKe-Book RCT study (Connor et al., 2019) (Table 2). Thus, while students made significant gains from pretest to posttest, there was not an effect of the strategy version for *Dragon's Lair*. We also tested an interaction between strategy version and WMG and L2M pretest scores to assess the extent to which gains depended on baseline decoding and comprehension skills and vocabulary knowledge. However, no significant interactions were found, indicating that the effect of each strategy version did not depend on baseline skills and knowledge. Finally, we also calculated how many times students read *Dragon's Lair*. On average, students read the story approximately two times (Table 1), however reading the story more than once was not associated with higher posttest outcomes. (Table 2).

*Hurricane!*. For *Hurricane!*, a paired samples *t*-test revealed that there was a significant difference between students' pretest and posttest scores on the *Hurricane!* Concept Knowledge Task ($t(252) = 14.785, p < .001$). We also ran a two-level HLM model with *Hurricane!* Concept Knowledge Task posttest scores as the outcome variable. Similar to the *Dragon's Lair* Word Knowledge Task analysis, grade, WMG (i.e., vocabulary), and L2M (i.e., decoding and comprehension), were added to the model. Both grade level and L2M were significant predictors of posttest scores in this model. To determine whether there were differences by strategy version of *Hurricane!* that students read, three dummy variables were created using the same method described above. Like *Dragon's Lair*, the summary and question generation versions of *Hurricane!* were not associated with higher posttest outcomes compared to the general comprehension strategy version of the book. Thus, while students made significant gains from pre to posttest, it did not necessarily matter which version of the book they read (Table 3). We also examined whether there was an interaction between students' L2M scores and strategy version: we found no significant effects for any of the three strategy versions. Thus, with both WKe-Books, students made significant gains from pretest to posttest, irrespective of strategy version or prior decoding and comprehension skills.

Higher scores on the embedded questions in *Hurricane!* also significantly predicted posttest scores on the *Hurricane!* Concept Knowledge posttest (Table 3). The better students did on the embedded questions, the better they did on the posttest. Finally, students who read *Hurricane!* two or more times also tended to have higher scores on the *Hurricane!* posttest.

**Research Question 2: What were teachers' perceptions of the WKe-Books as an instructional tool for improving comprehension skills?**

Overall, the feedback gathered during two focus groups with participating teachers on the WKe-Books was positive. The teachers ($n = 11$) reported that they thought the WKe-Books were a fun and engaging tool for their students. They reported that their students liked that they could name their own characters and choose their own adventure, and multiple students read the story more than once choosing different story streams, which was also evidenced in the user logs. One group also shared that they enjoyed how each book focused on one comprehension strategy which would be useful when working on specific strategies with students each week. One teacher said, "It [WKe-Books] would be great to have as a center activity over a period of time. Where they learn the type of questions that they would be answering. Allowing discussing amongst the group. More of a homogeneous grouping." Another said, "I actually enjoyed the different question types. It is excellent for the students to be able to see and be able to respond to these varying types of questions." In regards to *Hurricane!*, one teacher said, "My class loved it, it went together with the science unit benchmark. I can see why they [students] loved it so much – they got to pick their character names."

We also asked teachers to review and provide feedback on the lesson plans delivered by interventionists. Teachers indicated that the lesson plans were simple, easy to follow, and they could deliver them to their students. However, they also noted that small group reviews would likely be more effective for students than whole group instruction. For example, one teacher said, "it felt like the lesson plan was too small/short. A better lesson plan in place – some students need more support. Another shared, "small groups may be more efficient and effective." Another teacher echoed this feedback and added, "it was usually the same students participating so small groups may increase individual engagement." Multiple teachers also noted that WKe-Books could be difficult to implement with limited flexibility to introduce new tools or activities outside of their normal curriculum. For example, "Realistically, I would be limited with the amount of time that the students would be able to access the WKe-Book." Another said, "time constraints given our other requirements" would be their biggest challenge in implementing the WKe-Books into their curriculum and added, "but this might be good for intervention students with reading coaches."

Finally, teachers also shared that they found value in the feedback that was provided in the WKe-Books: "It [the feedback] was extremely effective. When they answered a question and saw the feedback, it really helped them to focus on what specific information was being asked." Another shared, "I liked that they were sent back to try to answer the questions again. It provided a more engaging reading experience than printed books."

## 4. Discussion

In this feasibility study, we examined two newly developed WKe-Books, one fiction and one hybrid science text, and their benefit on students' reading-related outcomes. There were two noteworthy findings. First, students who read the WKe-Books had improved posttest scores in word knowledge taught in *Dragon's Lair* and hurricane concept knowledge taught in *Hurricane!*. However, these gains did not appear to depend on the WKe-Book version students read. These results mirror results from the previous RCT on the initial version of the WKe-Book (Connor et al., 2019) that demonstrated the effectiveness of the WKe-Books on students' reading-related outcomes. For both books, we found an effect of grade level such that fourth and fifth graders tended to have higher posttest scores compared to third grade students.

We found no significant differential effects on posttest outcomes based on the strategy version students read for either WKe-Book. This was somewhat surprising for *Hurricane*! which was focused on learning concepts. In *Dragon's Lair*, the deeper text-meaning strategies did not appear to hinder word-meaning comprehension. While there were differences between these three strategies, ultimately, the aim of the strategies was the same - to teach difficult vocabulary and targeted concepts about hurricanes. For students who read *Hurricane!*, they were assigned to read a different strategy version than what they were assigned for *Dragon's Lair*, thus its possible effects of these strategies were minimized due to the fact that these students had been taught two of the three targeted strategies in the study. Although this study was not designed to be an RCT, taken together with the results of the RCT which included many of the same features (Connor et al, 2019), the results of this study demonstrate the potential utility of a tool like WKe-Books to build comprehension skills through both lower-level strategies to support word-meaning and text-meaning comprehension (i.e., rereading) and deeper-level strategies (i.e., summarization and question generation). These results also show the potential to expand e-book use beyond just literacy instruction to other content areas, such as science.

While we did observe significant pretest to posttest gains for *Dragon's Lair* in this study, we note that larger gains were observed in the previous RCT study for students who participated in weekly small group book clubs, particularly in the delayed treatment group ($M = 14.54$, $SD = 6.04$; Connor et al., 2019). However, our results also indicate that students in the current study who participated in whole class book clubs had larger gains ($M = 13.34$, $SD = 5.31$) than those in the previous RCT study who read the WKe-Book alone without small group book clubs ($M = 12.93$, $SD = 5.41$; Connor et al. 2019) demonstrating that additional review time outside of reading e-books alone, even if done in larger groups, can potentially yield greater learning outcomes. More focused individualized small groups book clubs may be most beneficial as students can be grouped by baseline reading skills, story streams, and/or progress in the book. Unlike the previous RCT study where students in a small group met with the interventionist once a week for 20 minutes, all students in this study had a 10-minute session with the interventionist three times a week (i.e., every time they read the WKe-Book), thus students in the current study received 30 minutes of review each week.. Taken together, results from these studies suggest that while e-books like the WKe-Books can be beneficial in improving students' learning outcomes on their own, effects can be greater with additional support from teachers to reinforce concepts and vocabulary. For feasibility of implementation, if teachers are limited with time to run small group book clubs, it is possible that a larger group format can also be effective. More studies will be needed to better understand the impact on literacy outcomes with book clubs varying in format (whole class or small group) and the frequency in which they are implemented. Future research may also wish to assess the feasibility and effectiveness of implementing some of the aspects of the book club reviews directly into the WKe-Books.  Developing new ways to include additional features that review important vocabulary and concepts more directly in e-books (such as end-of-chapter reviews) may yield greater outcomes in addition to the support a teacher can provide.

    While students were only asked to read each book one time, if they finished early, they were encouraged to read again and try selecting different story paths. Reading *Hurricane!* more

than once was associated with greater posttest gains, however this was not the case in *Dragon's Lair*. In *Hurricane!,* we ensured that regardless of story stream, the same targeted concepts were taught, thus repeated readings allowed students to review the concepts again. In *Dragon's Lair*, not all 25 words were taught across every stream. On average, students were exposed to about 13 of the words, thus, some students may have only had very minimal exposure to some of the vocabulary through the opening review sessions. Secondary readings were counted based on the number of times students started the books again, but it did not necessarily mean that they fully completed a second reading, especially in *Dragon's Lair*, which was longer. Future books utilizing a choose-your-own-adventure format should ensure streams are balanced in targeted words, concepts, and embedded questions to allow for more accurate interpretation of posttest outcomes.

Embedded questions provided an opportunity for explicit instructional support prompting students to employ targeted comprehension strategies when they provided incorrect responses. Progress through the books was halted until students answered correctly, this encouraged students to stop and think carefully about the text they had just read. It is also possible that if students lost focus while reading the book and were just clicking through pages, the embedded questions served as a cue to re-focus their attention on the story.

The results of this study align with Furenes et al. (2021) finding that e-book features that enhance the content of the book may be most beneficial for students. The features implemented in the WKe-Books were focused on supporting students' comprehension of the text. In both books, some decisions led to dead ends in the story, which encouraged students to read carefully and make appropriate decisions to avoid being sent backwards in the book. The interactive features of the WKe-Books were also focused on enhancing students' engagement by allowing them to name their characters after themselves or friends, providing a more personal connection to the stories. On average, for both WKe-Books, students read each story two or more times as they wanted to see what would happen if they picked a different path.

**Feasibility**

Teacher feedback on the feasibility of using the WKe-Books in classrooms provided helpful insights. First, they reported that the students enjoyed the interactive and personalized nature of the books and that they would consider using them again in the future. Multiple teachers agreed that while the whole class review sessions with interventionists were helpful, the lesson reviews would likely have been more effective if done in small groups. This feedback is aligned with results from the initial pilot study in which book clubs were both longer in length (30 minutes) and organized in small groups of 5-6 students (Connor et al., 2019). Importantly, teachers reported the biggest hurdle in implementing a tool like the WKe-Books is lack of time outside of their regular curriculum. A larger future goal of this work might be to partner in the research and development of e-books more directly with curriculum developers so that implementation of such tools are aligned with daily curriculum.

**Implications for Design of Digital Books**

Provision of immediate feedback to multiple-choice items appeared to be beneficial for students in this study. Knowing that they answered a question incorrectly, in real time, provided them with an opportunity to go back and employ the targeted comprehension strategies to repair their understanding of complex text. The WKe-Book design did not allow students to proceed until they chose the correct answer. Future work may also consider how advances in artificial intelligence (AI) and natural language processing (NLP) technology can be leveraged to provide more detailed and individualized feedback to students while reading.

The affordances of digital books, such as the WKe-Books, can also provide educators and parents with feedback on students' performance. In this way, digital books can serve as a stealth assessment of reading skills (Shute, 2011) and offer insight into reading behaviors, something that cannot be obtained as easily through printed books. We used log data to examine chosen story paths, how many attempts were necessary to answer questions, and how many times they read the whole story. Designing user-friendly reporting interfaces for teachers can provide helpful feedback and progress-monitoring on reading performance.

**Limitations**

A major limitation during the study was the abrupt closure of schools due to the COVID-19 pandemic. This prevented us from administering posttests for the L2M and WMG assessments and a comprehension monitoring task to assess students' calibration, strategy use, and potential transfer of the strategies students learned in the WKe-Books. Additionally, our AZ partner school was not able to read *Hurricane!* resulting in a smaller sample size to address research questions related to *Hurricane!*. Future studies should also consider implementing a randomized control trial to better assess the effectiveness of additional strategies as was done in the initial testing of *Dragon's Lair* (Connor et al. 2019).

Finally, while it was beyond the scope and resources available in the current study, analysis of the open-ended responses in the book may provide greater insight into the effectiveness of the strategies. Since the *Hurricane!* Concept Knowledge Task only included one question generation and one summary question, this made it more difficult to assess whether students' summaries and questions improved as they read the books. Future work may wish to consider building pre and post assessments that include more question generation and summary-type questions in order to better assess the effectiveness of the strategies taught in the WKe-Books.

**Conclusions**

The interactive design of the WKe-Books–presenting students with immediate performance feedback, allowing students to name their characters, and providing agency to choose different story streams–offered an engaging and impactful learning experience. Overall, the findings from this study suggest features and strategies to improve the design of e-books that show promise in enhancing word learning and reading comprehension. This study contributes to the literature by providing insights into the affordances of e-books that can effectively support reading comprehension outcomes for students.

**Tables**

Table 1. *Dragon's Lair and Hurricane! Descriptive Statistics*

|  | *N* | Mean (SD) | Minimum | Maximum |
|---|---|---|---|---|
| Word Match Game Growth Scaled Score | 421 | 490.53 (13.91) | 457.00 | 678.07 |
| Letters2Meaning Growth Scaled Score | 416 | 650.81 (83.33) | 312.32 | 926.09 |
| Dragon's Lair Word Knowledge Pretest | 421 | 12.21 (4.63) | 1 | 28 |
| Dragon's Lair Word Knowledge Posttest | 421 | 13.34 (5.31) | 0 | 29 |
| Dragon's Lair Times Story was Read | 419 | 2.20 (.806) | 1 | 3 |
| Hurricane! Concept Knowledge Pretest | 271 | 8.63 (2.45) | 1 | 14 |
| Hurricane! Concept Knowledge Posttest | 258 | 11.07 (3.02) | 1 | 18 |
| Hurricane! Embedded Question Score | 255 | 24.47 (6.21) | 0 | 36 |
| Hurricane! Times Story was Read | 255 | 2.11 (.816) | 1 | 3 |

*Table 2. Hierarchical Linear Model Predicting Dragon's Lair Posttest Gains*

| **Fixed Effects** | **Coefficient** | **Standard Error** | *t*-ratio | *d.f.* | *p*-value |
|---|---|---|---|---|---|
| Intercept, β0 | | | | | |
|     Fitted Mean, γ00 | 13.133 | 0.162 | 80.861 | 14 | <.001 |
|     Grade Level, γ01 | 0.562 | 0.226 | 2.487 | 14 | 0.026 |
| For slope, β1 | | | | | |
|     Word Match Game Growth Scaled Score, γ10 | 0.025 | 0.013 | 1.964 | 369 | 0.050 |
| For slope, β2 | | | | | |
|     Letters2Meaning Growth Scaled Score, γ20 | 0.021 | 0.003 | 8.138 | 369 | <.001 |
| For slope, β3 | | | | | |
|     Dragon's Lair Pretest Total Score, γ30 | 0.559 | 0.047 | 11.941 | 369 | <.001 |
| For slope, β4 | | | | | |
|     Dragon's Lair Question Generation Strategy, γ40 | -0.134 | 0.433 | -0.311 | 369 | 0.756 |
| For slope, β5 | | | | | |
|     Dragon's Lair Summary Strategy, γ50 | 0.381 | 0.394 | 0.966 | 369 | 0.335 |
| For slope, β6 | | | | | |
|     Dragon's Lair Total Times Read, γ60 | 0.245 | 0.206 | 1.187 | 369 | 0.236 |

Final estimation of variance components:

| **Random Effect** | **SD** | **Variance Component** | *d.f.* | Chi-square | *p*-value |
|---|---|---|---|---|---|
| Intercept, *u*0 | 0.429 | 0.185 | 9 | 17.92 | 0.036 |
| level-1. *r* | 2.208 | 4.876 | | | |

Deviance= 2015.159

*Note. Grade is centered at fourth grade. Word learning is the reference strategy for strategy version.*

Table 3. *Hierarchical Linear Model Predicting Hurricane! Posttest Gains*

| Fixed Effects | Coefficient | Standard Error | *t*-ratio | *d.f.* | *p*-value |
|---|---|---|---|---|---|
| Intercept, $\beta_0$ | | | | | |
|    Fitted Mean, $\gamma_{00}$ | 10.832 | 0.352 | 30.73 | 8 | <.001 |
|    Grade Level, $\gamma_{01}$ | 0.581 | 0.297 | 1.956 | 8 | 0.051 |
| For slope, $\beta_1$ | | | | | |
|    Word Match Game Growth Scaled Score, $\gamma_{10}$ | 0.013 | 0.019 | .661 | 167 | 0.509 |
| For slope, $\beta_2$ | | | | | |
|    Letters2Meaning Growth Scaled Score, $\gamma_{20}$ | 0.007 | 0.003 | 2.447 | 167 | 0.015 |
| For slope, $\beta_3$ | | | | | |
|    Hurricane! Pretest Total Score, $\gamma_{30}$ | 0.497 | 0.083 | 5.960 | 167 | <.001 |
| For slope, $\beta_4$ | | | | | |
|    Hurricane! Question Generation Strategy, $\gamma_{40}$ | 0.474 | 0.493 | 0.961 | 167 | 0.338 |
| For slope, $\beta_5$ | | | | | |
|    Hurricane! Summary Strategy, $\gamma_{50}$ | 0.504 | 0.500 | 1.008 | 167 | 0.315 |
| For slope, $\beta_6$ | | | | | |
|    Hurricane! Embedded Questions % Score, $\gamma_{60}$ | 5.552 | 1.486 | 5.563 | 167 | <.001 |
| For slope, $\beta_7$ | | | | | |
|    Hurricane Total Times Read, $\gamma_{70}$ | 0.459 | 0.217 | 2.120 | 167 | 0.035 |

Final estimation of variance components:

| Random Effect | SD | Variance Component | *d.f.* | Chi-square | *p*-value |
|---|---|---|---|---|---|
| Intercept, $u_0$ | 0.285 | 0.081 | 8 | 11.487 | 0.175 |

| | | |
|---|---|---|
| level 1 | 2.267 | 5.142 |

Deviance= 825.992

*Note.* Grade is centered at fourth grade; General comprehension is the reference group for strategy version.

INTERACTIVE E-BOOKS AND COMPREHENSIONKorat, O., Tourgeman, M., & Segal-Drori, O. (2022) E-book reading in kindergarten and story comprehension support. *Reading and Writing, 35,* 155–175. https://doi.org/10.1007/s11145-021-10175-0

López-Escribano, C., Valverde-Montesino, S., & García-Ortega, V. (2021). The impact of e-book reading on young children's emergent literacy skills: An analytical review. *International Journal of Environmental Research and Public Health*, *18*(12), 6510. https://doi.org/10.3390/ijerph18126510

McDonald, A.-E. (2012). *The dragon's lair: The story of the Scarlett Square.* Captiva Island, FL: Beach Walk Books.

McNamara, D. S. (Ed.). (2007). *Reading comprehension strategies: Theory, interventions, and technologies.* Mahwah, NJ: Erlbaum.

McNamara, D. S., Ozuru, Y., & Floyd, R. G. (2011). Comprehension challenges in the fourth grade: The roles of text cohesion, text genre, and readers' prior knowledge. International *Electronic Journal of Elementary Education, 4*(1), 229-257.

Raudenbush, S. W., & Bryk, A. S. (2002). *Hierarchical linear models: Applications and data analysis methods* (second ed.). Thousand Oaks, CA: Sage.

Reich, S., Yau, J. C., & Warschauer, M. (2016). Tablet-based ebooks for young children: What does the research say? *Journal of Developmental and Behavioral Pediatrics, 37*(7), 585–591. https://doi.org/10.1097/DBP.0000000000000335

Rosenshine, B., Meister, C., & Chapman, S. (1996). Teaching students to generate questions: A review of the intervention studies. *Review of Educational Research, 66*(2), 181-221.

# Appendix

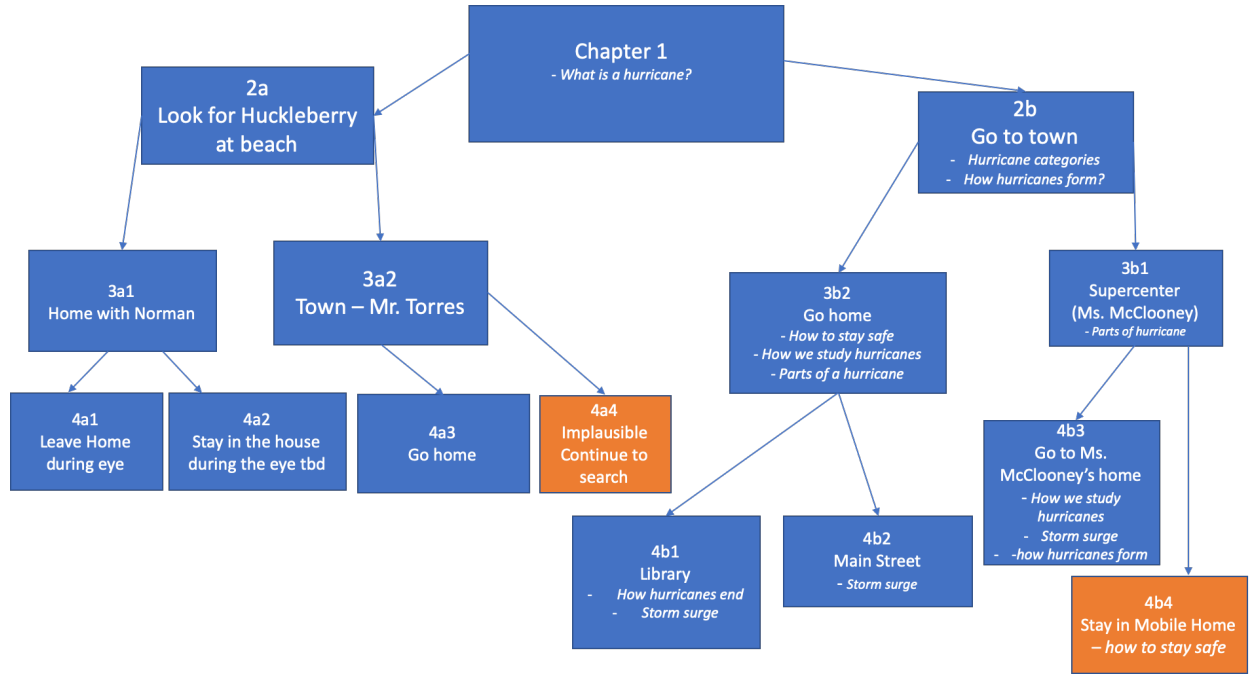

*Hurricane Story Structure.*

## Hurricane Sample Illustrations

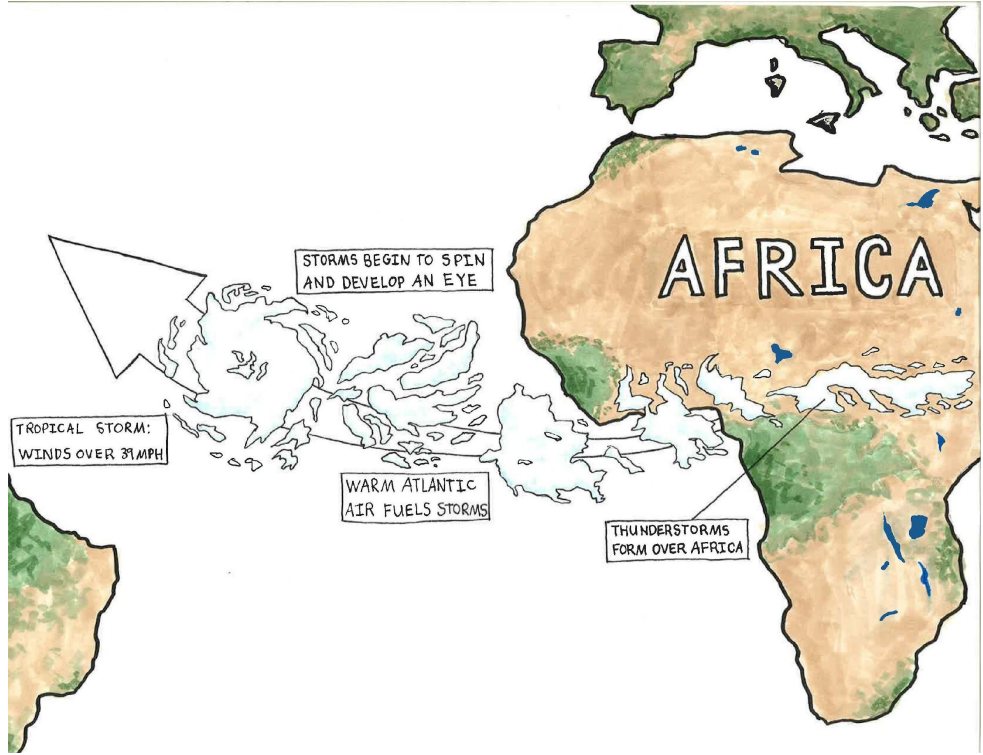

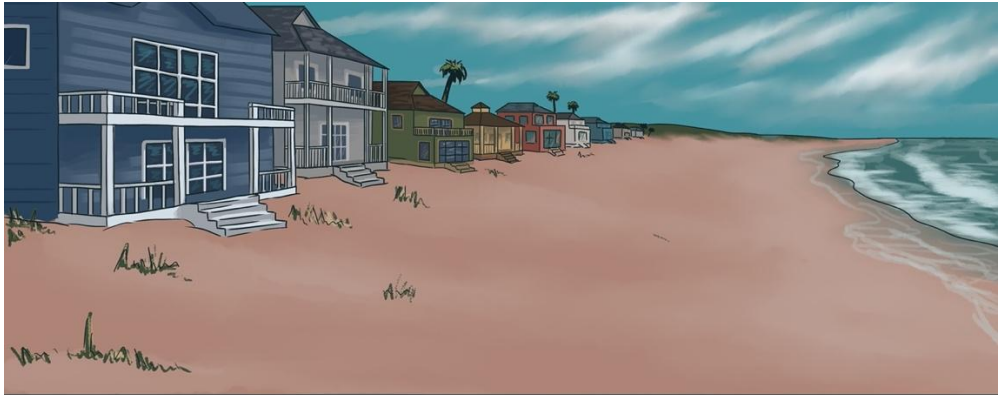

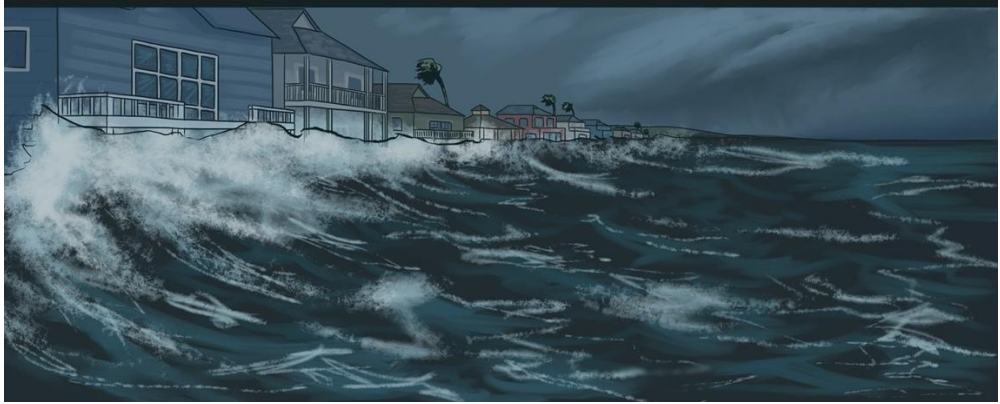

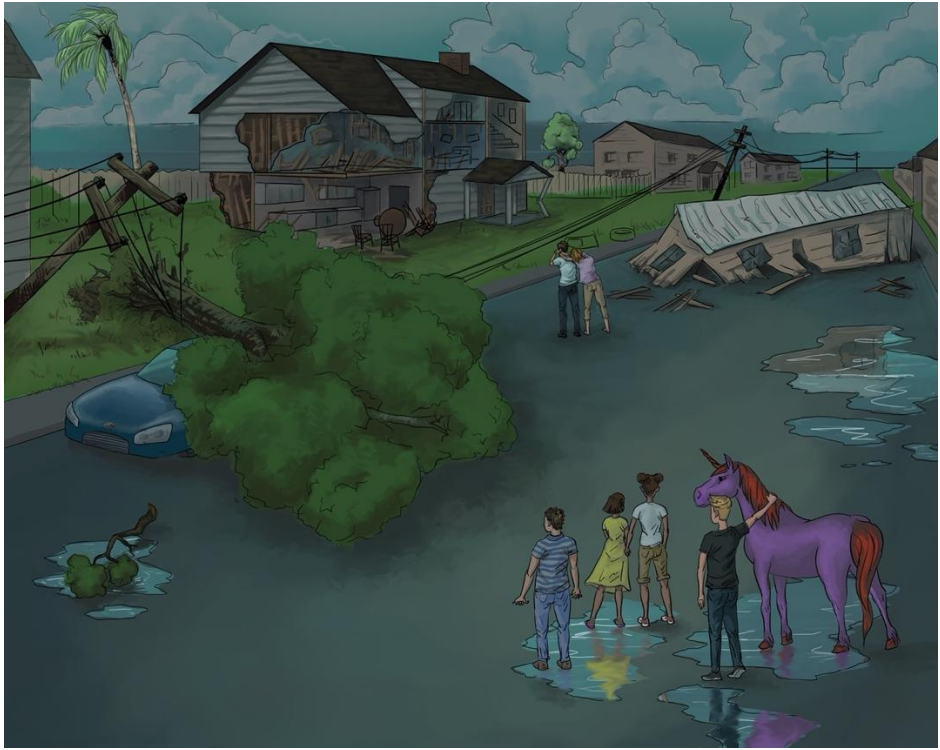

*Note.* Illustrations contributed by Brady Fellows

## WKe-Book Sample Embedded Questions

*Dragon's Lair* **General Comprehension Question and Feedback:**

## What does Katie mean when she asks whether she and Charlie were impulsive?

A. She wonders whether they made a quick and decisive decision to run.
B. She wonders whether they acted without thinking through the evidence and consequences.
C. She wonders whether they were stupid to run away.
D. She wonders whether they were smart to run away.

## Impulsive General Comprehension Feedback

Correct answer feedback:

"You are right! Impulsive means to act on impulse and impulses include feelings, like fear, and sudden urges. And these urges can be involuntary. That means that you can't resist them. Impulses can have positive connotations as well, such as a generous impulse. History of the word: Impulse is from the Latin word, impulsis, which means pressure. This is also where the word pulse comes from. Keep reading!"

Incorrect answer feedback:
"You are not quite right! Impulsive means Impulsive means to act on impulse and impulses include feelings, like fear, and sudden urges. And these urges can be involuntary. That means that you can't resist them. It has nothing to do with being smart or stupid. History of the word: Impulse is from the Latin word, 'impulsis', which means pressure. This is also where the word pulse comes from. Keep reading now that you know what impulsive means!"

*Dragon's Lair* **Summary Question and Feedback:**

## Which description below provides best summarizes what happens after Charlie, Katie, and Finn find the red dragon again?

A. The red dragon tells them how the green dragons are the ones that raid the town and steal children, not the red dragons. They need children to work in the caves to find emeralds because the dragons are too large. The green dragons are planning to steal all of the children in the town next. King Edward cannot stop them because he has lost all of his power. The dragons bring them chicken, sausages, and milk to eat. The green dragons have stolen the magical emerald and the Golden Eagle.

B. King Edward tells them that the other dragons have stolen the magical emerald and plan to kidnap all of the children to work in the mines.

C. King Edward tells them that while the red dragons are benevolent, the green dragons are malevolent and plan to kidnap all of the children in the town to work in the mines. The green dragons have stolen the emerald and the Golden Eagle and plan to use the emerald to do evil things. The red dragons are too large to fit in the caves to rescue the eagle and retrieve the emerald.

D. King Edward tells them that the green dragons have kidnapped the Golden Eagle and stolen the magical emerald and plan to use it to help the town. The red dragons are too large to fit into the caves to get the emerald back.

## Red Dragon Summary Question Feedback

A. You are not quite right. This description is too long and contains details that are not relevant to the overall story. For example, it is not an important detail that that the kids were fed chicken and sausages. Try reading the question and choices again to see if there might be a better summary.

B. You are not quite right. This description is missing important details. For example, an important detail of the story is that the red dragons are benevolent while the green dragons are malevolent. Try rereading the page and other choices again to see if there might be a better summary.

C. You are right. This summary provides all of the important details from what the red dragon has described about the green dragons, the eagle, and the emerald. Keep reading!

D. You are not quite right. The green dragons are malevolent and plan to use the emerald to do bad things. Try reading the last page again and the choices to see if there is a better summary what just happened in the story.

*Dragon's Lair* **Question Generation and Feedback:**

## Which question below is the *best question* to ask yourself right now?

A. Why did Katie and Charlie act impulsively by running away from the red dragon?
B. How did Katie and Charlie save the Ryan twins?
C. Why did Katie and Charlie cogitate before running away from the red dragon?
D. Are Charlie and Katie lying about coming back to save the Ryan twins?

## Impulsive Question Generation Feedback

A. You are right! Why do you think Katie and Charlie ran away from the red dragon? What does Katie mean when she asks Finn if she and Charlie were impulsive? Keep reading!

B. You are not quite right. Charlie and Katie have not yet saved the Ryan twins. They couldn't get them out of their cell, but they have promised to come back for them. Try reading the page again to see if there might be a better question to ask yourself right now.

C. You are not quite right. Katie and Charlie did not *cogitate* before running from the red dragon. *The prefix 'cog' means 'to think', to recognize."* What do you think 'cogitate' means then? Katie asks if they were 'impulsive' by running away from the red dragon? Do impulsive and cogitate mean the same thing? Try reading the page again to see if there might be a better question to ask yourself right now.

D. You are not quite right. Charlie and Katie want to help save the Ryan twins. They are not working with the evil green dragons or Doc Wilson. Try reading the page again to see if there might a better question you could ask yourself right now.

*Hurricane!* **General Comprehension Question and Feedback:**

## What is the most dangerous part of a hurricane and why is it so dangerous?

A. High winds, they can get up to 157 mph and destroy many unstable buildings.
B. Rain, this can cause massive flooding in areas affected by hurricanes.
C. Atmospheric pressure, this can cause major structural damage to buildings.
D. Storm surge, this can cause significant flooding and damage.

## Storm Surge General Comprehension Question Feedback

1. You are not quite right. Although these high winds are very dangerous, there is another part of a hurricane that is more dangerous. Try again!
2. You are not quite right. Although rain can cause massive flooding, most areas affected by hurricanes are prepared for the flooding rain causes. Try again!
3. You are not quite right. There is another part of a hurricane that is the most dangerous. Try again!
4. You are right! Storm surges cause significant flooding that can go over 20 feet high. The surge can come in for miles inland, cutting off roads and building, trapping people. A storm surge is the leading cause of death during a hurricane. Keep reading!

*Hurricane!* **Summary Question and Feedback:**

## Which of the descriptions below provides the best summary about storm surge?

A. Storm surge occurs when hurricane winds cause the ocean waters to rise above the normal high tide causing significant flooding and damage. Some storms can cause storm surge of over 20 feet high depending on the intensity of the storm and the geography of the coastline. The surge can come in for miles inland, cutting off roads and buildings, trapping people. Thus, storm surge is the leading cause of death in a hurricane.

B. Storm surge occurs when hurricane winds cause the ocean waters to rise above the normal high tide causing significant flooding and damage. Storm surge is also the leading cause of death in a hurricane.

C. Storm surge is the least dangerous part of a hurricane. It brings high winds but does not cause significant damage.

D. Storm surge occurs when hurricane winds cause significant flooding and damage.

## Storm Surge Summary Question Feedback

1. You are not quite right. This summary includes the main ideas, such as what a hurricane is and the most dangerous parts, but it also has too much information. For example, knowing that a storm surge can reach over 20 feet is small detail and not a main idea. Try reading each choice again to see if there is a better summary!

2. You are right! This summary includes the most important details about storm surge. Keep reading!

3. You are not quite right. This summary does not provide accurate information about storm surges. A storm surge is the most dangerous part of a hurricane, it brings the most damage and even deaths. Try reading each choice again to see if there is a better summary!

4. You are not quite right. This summary is missing some important details about storm surges. Think about what other details should be added to summarize storm surges. Try reading each choice again to see if there is a better summary!

*Hurricane!* **Question Generation Question and Feedback**:

## Which question below is the best question you could ask yourself in order to check for your understanding of storm surge?

A. When will Hurricane Howard be over?
B. Why is storm surge the least dangerous part of a hurricane?
C. What is storm surge and why is it so dangerous?
D. Will Dr. Brown keep the kids safe?

## Storm Surge Question Generation Feedback

A. You are not quite right. Although the it would be nice to know when the hurricane will end, this question won't really help you best understand the information that you just read. Try reading the choices again and see if there might be a better question you could ask yourself right now.

B. You are not quite right. Dr. Brown has told the children that storm surge is actually the most dangerous threat from hurricanes. Try reading the choices again and see if there might be a better question that you could ask yourself.

C. You are right! Based on the information presented in the text, storm surges can cause a lot of damage because the winds create a giant wall of water to rise above normal sea levels. This can flood the island. What is storm surge and why is it so dangerous? Keep on reading!

D. You are not quite right. Dr. Brown is trying to convince the kids to go back into the safety of their home. This question won't help you best understand the information that you just read. Try reading the choices again and see if there might be a better question you could ask yourself right now.